# Rail Vehicle Localization and Mapping with LiDAR-Vision-Inertial-GNSS Fusion

Yusheng Wang, *Graduate Student Member, IEEE*, Weiwei Song, Yidong Lou, Yi Zhang, Fei Huang, Zhiyong Tu, and Qiangsheng Liang

*Abstract*—In this paper, we present a global navigation satellite system (GNSS) aided LiDAR-visual-inertial scheme, RailLoMer-V, for accurate and robust rail vehicle localization and mapping. RailLoMer-V is formulated atop a factor graph and consists of two subsystems: an odometer assisted LiDAR-inertial system (OLIS) and an odometer integrated Visual-inertial system (OVIS). Both the subsystem exploits the typical geometry structure on the railroads. The plane constraints from extracted rail tracks are used to complement the rotation and vertical errors in OLIS. Besides, the line features and vanishing points are leveraged to constrain rotation drifts in OVIS. The proposed framework is extensively evaluated on datasets over 800 km, gathered for more than a year on both general-speed and high-speed railways, day and night. Taking advantage of the tightly-coupled integration of all measurements from individual sensors, our framework is accurate to long-during tasks and robust enough to grievously degenerated scenarios (railway tunnels). In addition, the real-time performance can be achieved with an onboard computer.

*Index Terms*—train, multi-sensor, localization and mapping,

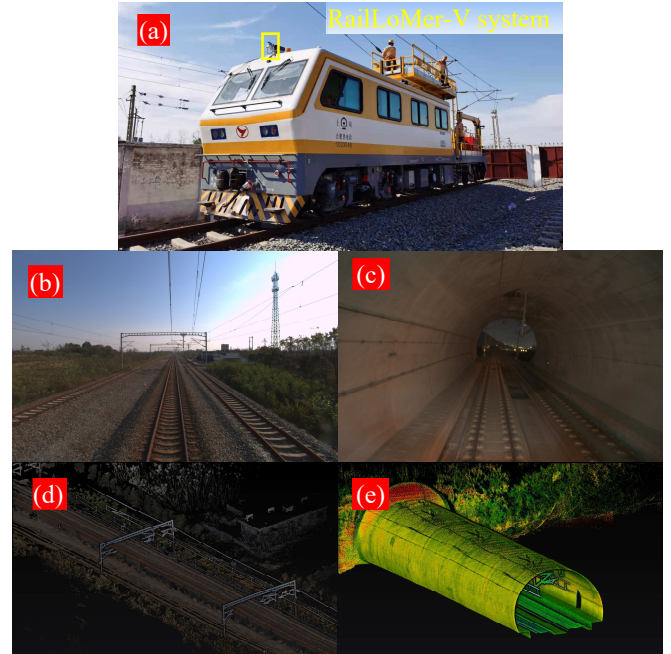

**Fig. 1.** Overview of the proposed RailLoMer-V system. (a) is an example of maintenance vehicle equipped with RailLoMer-V system. (b) and (c) denotes two typical environments, with (b) the day view of general-speed railway and (c) is the night view of high-speed railway tunnel. (d) and (e) is the respective colored-mapping and LiDAR mapping result of (b) and (c).

## I. INTRODUCTION

PRECISE rail vehicle localization and long-term railroad environment monitoring is of critical importance for safety operation on railway. The current positioning strategy is still dominated by trackside systems, which not only lack efficiency and accuracy for real-time applications, but also require large civil investment for construction and successive maintenance. In addition, the power supply of typical section needs to be cut off for infrastructure maintenance, and the trackside systems work not properly for high-speed railway maintenance vehicles.

Recent advances in sensor technology and railway signal standardization have prompted research of onboard sensors. With the flexibility of installation and accurate positioning in open districts, the global navigation satellite system (GNSS) have attracted many researchers. Besides, the track geometry, IMU, odometer can supplement the system at GNSS outages. However, these approaches merely acquire the train state data without consideration of the environmental information. To cope with this problem, integration with perception sensors, is also required, and many previous works have employed mobile mapping system (MMS). As a direct-georeferencing method, MMS-based solution requires a series of post processing and survey-grade instruments. Therefore, they are not satisfactory for real-time rail vehicle location and large deployment.

Different from commonly used unmanned aerial vehicles (UAV) and unmanned ground vehicles (UGV), researchers are pretty open in choosing localization sensors and computation units for rail applications. With the framework of estimating train state and mapping the surrounding in the meantime, simultaneously localization and mapping (SLAM) is a promising approach towards rail vehicle localization and environment monitoring problems. However, a number of difficulties affect the application of SLAM on rail vehicles.

1) *Long-Time Constrained Motions*: the rail vehicles are constrained to move along planar trajectories, leading to spurious information gain to the unobservable directions for IMU biases. This potential observability issue will result in large scale drift for many Visual-inertial approaches.

2) *Repetitive Features*: shown in Fig. 1 (b) and (c), the mainly observable features are the repetitive rail tracks and suspension clamps, which is challenging for feature tracking based methods.

3) *No Revisited Places*: many SLAM approaches employ the place descriptor to detect revisited places, and correct the accumulated drifts atop the detected loops. However, there are no revisited places for trains, which put forward higher requirements for low drift pose estimation.

To tackle these challenges, we propose RailLoMer-V, an accurate and robust system for rail vehicle localization and mapping. RailLoMer-V fuses multi-modal information with a tightly-coupled manner. Our design of RailLoMer presents the following contributions:

1) We propose a framework that tightly fuses LiDAR, IMU, rail vehicle wheel odometer, camera, and GNSS through sliding window based factor graph formulation.
2) We fully leverage the geometric information of sensor measurements, where plane constraints from extracted rail tracks and vanishing point are utilized to increase system accuracy and robustness.
3) Our framework is evaluated with data gathered over a year, covering various scales, weathers, and railways.

To the best of the authors' knowledge, RailLoMer-V is the first solution to real-time and large-scale rail vehicle SLAM. Some of the mapping results are shown in Fig. 1, denoting the capability of handling diverse railroad scenarios.

## II. RELATED WORK

Prior works on train localization methods and LiDAR-visual-inertial SLAM are extensive. In this section, we briefly review works on train localization and LiDAR-visual-inertial SLAM.

### A. Train Positioning Solutions

The most common train positioning system is based on trackside sensors, such as a Balise [1], [2]. This system divides the railroad into individual blocks, where a Balise is placed at the beginning of each block. When a train passes over it, the Automatic Train Control (ATC) system detects and locates that a train is within that block. Considering its large capital investment and low localization efficiency, many researchers seek to supplement the limitations with either onboard sensors or feature matching based methods.

The onboard sensors include radio frequency identification (RFID) [3], Doppler radar [4], GNSS receivers [5], IMU [6], and the tachometer or odometer [7]. The feature-matching methods first establish the feature database, then the real-time positions can be acquired through matching with the database. Such database consists of track signature, digital track line, or laser scan features [8].

The potential of SLAM for rail vehicles localization and mapping has not been well investigated. One of the early works, RailSLAM, jointly estimated the train state and validated the correctness of initial track map based on a general Bayesian theory [9]. The performance of Visual-inertial odometry on rail vehicles have been extensively evaluated in [10], [11], indicating that the Visual-inertial odometry is not reliable for railroad applications. But the LiDAR-visual-inertial based SLAM is still an open problem for railway applications.

### B. LiDAR-Visual-Inertial SLAM

To increase system robustness against sensor failures, fusion of multi-modal sensing capabilities has been explored in many scholarly works. These schemes can be categorized into either loosely-coupled [12], [13] or tightly-coupled approaches [14]–[16]. And the geometric structures can be employed to further avoid degeneracy at self-repetitive districts. Huang *et.al.* extract point and line features from images, and use LiDAR to assign depth information as well as scale correction [17]. Compared with point-only approaches, their method greatly reduces the feature ambiguity and achieves accurate estimation accuracy. The detected line segments can be also used to find vanishing point [18] for additional rotation constraint. Wisth *et.al.* propose to utilize line and planar primitives from LiDAR and track them over multiple scans [19]. The lightweight expression of the features allows for real-time execution on limited computation resource.

From the discussion of the literature, we can see that the rail vehicle SLAM has not been well solved and evaluated. And this paper seeks to achieve real-time, low-drift and robust odometry and mapping for large-scale railroad environments with geometric structure assisted LiDAR-visual-inertial SLAM.

## III. PROBLEM STATEMENT

We seek to estimate the trajectory and map the surrounding of a rail vehicle with multi-sensor measurements, in which the state estimation procedure can be formulated as a maximum-a-posterior (MAP) problem.

We denote $(\cdot)^W$, $(\cdot)^B$, $(\cdot)^C$ and $(\cdot)^O$ as the world, body, camera and odometer frame. In addition, we define $(\cdot)^B_W$ as the transform from world frame to the IMU frame. We use both rotation matrix $\mathbf{R}$ and quaternion $\mathbf{q}$ to represent rotation. Besides, we denote $\otimes$ as the multiplication between two quaternions. $(\hat{\cdot})$ is denoted as the estimation of a certain quantity.

### A. State Definition

The train state at time $t_i$ is defined as follows:

$$x_i = [\mathbf{p}_i, \mathbf{v}_i, \mathbf{q}_i, \mathbf{b}_a, \mathbf{b}_g, \mathbf{c}_i] \qquad (1)$$

where $\mathbf{p} \in \mathbb{R}^3$, $\mathbf{v} \in \mathbb{R}^3$, and $\mathbf{q} \in SO(3)$ are the position, linear velocity, and orientation vector. $\mathbf{b}_a$ and $\mathbf{b}_g$ are the usual IMU gyroscope and accelerometer biases. And $\mathbf{c}$ is the scale factor of the odometer.

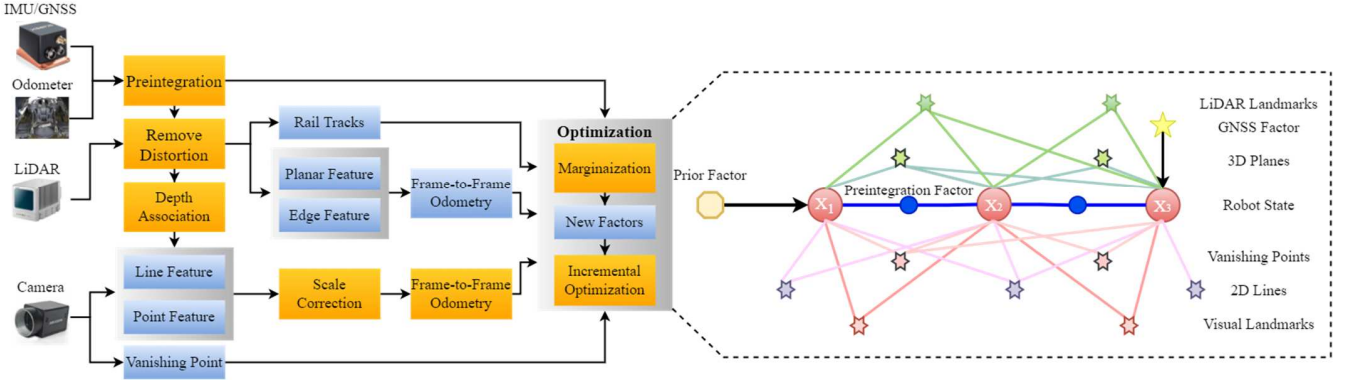

**Fig. 2.** The system framework of our approach, with the graph structure illustrated in the dashed box. Given the input LiDAR point cloud and monocular image, we extract the point, line, and vanishing point for each image, and track image frames with scale corrected by assigned depth information. Besides, the rail tracks are detected and extracted, with the rail track planes as plane constraints. All the measurements are jointly optimized with a sliding window based graph formulation.

*B. Maximum-a-Posterior Problem*

Given the measurements $\mathcal{Z}_k$ and the history of states $\chi_k$, the MAP problem can be formulated as:

$$\chi_k^* = \underset{\chi_k}{\mathrm{argmax}}\, p(\chi_k|\mathcal{Z}_k) \propto p(\chi_0)p((\mathcal{Z}_k|\chi_k)) \quad (2)$$

If the measurements are conditionally independent, then (2) can be solved through least squares minimization:

$$\chi^* = \underset{\chi_k}{\mathrm{argmin}} \sum \sum_{i=1}^{k} \|r_i\|^2 \quad (3)$$

where $r_i$ is the residual of the error between the predicted and measured value.

*C. Sliding Window Based Optimization*

To ensure the real-time performance of the optimization scheme, we exploit keyframes to establish sliding windows. For a sliding window of $N_K$ keyframes, the optimal states are obtained through minimizing:

$$\underset{\chi}{\min}\{\|r_\mathcal{P}\|^2 + \sum_{i=1}^{N_K}\|r_{\mathcal{J}_i}\|^2 + \sum_{i=1}^{N_{\mathcal{L}_K}} r_{\mathcal{L}_i} + \sum_{i=1}^{N_{\mathcal{P}_K}}\|r_{\mathcal{P}_i}\|^2 + \sum_{i=1}^{N_{\mathcal{M}_K}}\|r_{\mathcal{M}_i}\|^2 + \sum_{i=1}^{N_{\mathcal{S}_K}}\|r_{\mathcal{S}_i}\|^2 + \sum_{i=1}^{N_{\mathcal{V}_K}}\|r_{\mathcal{V}_i}\|^2 + \sum_{i=1}^{N_{\mathcal{G}_K}}\|r_{\mathcal{G}_i}\|^2\} \quad (4)$$

where $r_\mathcal{P}$ is the prior factor marginalized by Schur-complement [20], $r_{\mathcal{J}_i}$ is the residual of IMU/odometer preintegration result. $r_{\mathcal{L}_i}$ and $r_{\mathcal{P}_i}$ define the residual of LiDAR and plane constraints. $r_{\mathcal{M}_i}$, $r_{\mathcal{S}_i}$, and $r_{\mathcal{V}_i}$ denote the residual of reprojection, line segment, and vanishing point terms. The residual of global positioning system is $r_{\mathcal{G}_i}$.

IV. METHODOLOGY

The system overview and the constructed graph is shown in Fig. 2, we now describe the measurements and residuals of the factors in detail.

*A. IMU/Odometer Preintegration Factor*

The raw accelerometer and gyroscope measurements, $\hat{\mathbf{a}}$ and $\hat{\boldsymbol{\omega}}$, are given by:

$$\hat{\mathbf{a}}_k = \mathbf{a}_k + \mathbf{R}_W^{B_k}\boldsymbol{g}^W + \mathbf{b}_{a_k} + \boldsymbol{\eta}_a,$$
$$\hat{\boldsymbol{\omega}}_k = \boldsymbol{\omega}_k + \mathbf{b}_{\omega_k} + \boldsymbol{\eta}_\omega \quad (5)$$

where $\boldsymbol{\eta}_a$ and $\boldsymbol{\eta}_\omega$ are the zero-mean white Gaussian noise, with $\boldsymbol{\eta}_a \sim \mathcal{N}(\mathbf{0}, \boldsymbol{\sigma}_a^2)$, $\boldsymbol{\eta}_\omega \sim \mathcal{N}(\mathbf{0}, \boldsymbol{\sigma}_\omega^2)$. The gravity vector in the world frame is denoted as $\boldsymbol{g}^W = [0,0,g]^T$. And the model of odometer sensor is given by:

$$\boldsymbol{c}^{O_k}\hat{\mathbf{v}}^O = \mathbf{v}^O + \boldsymbol{\eta}_{s^O} \quad (6)$$

where $\boldsymbol{c}^{O_k}$ denotes the scale factor of the odometer modeled as random walk, with $\boldsymbol{\eta}_{s^O} \sim \mathcal{N}(\mathbf{0}, \boldsymbol{\sigma}_{s^O}^2)$. Then the pose estimation can be achieved through synchronously collected gyroscope and odometer output, and the displacement within two consecutive frames $k$ and $k+1$ can be given as:

$$\hat{\mathbf{p}}_{O_k}^{O_{k+1}} = \mathbf{p}_{O_k}^{O_{k+1}} + \boldsymbol{\eta}_{\mathbf{p}^O} \quad (7)$$

where $\boldsymbol{\eta}_{\mathbf{p}^O}$ is also the zero-mean white Gaussian noise. Based thereupon and the preintegration form in [20], we can formulate the discrete form of preintegrated IMU/odometer information between $k$ and $k+1$ $\left[\hat{\boldsymbol{\alpha}}_{B_{i+1}}^{B_k}, \hat{\boldsymbol{\beta}}_{B_{i+1}}^{B_k}, \hat{\boldsymbol{\gamma}}_{B_{i+1}}^{B_k}, \hat{\boldsymbol{\phi}}_{B_{i+1}}^{B_k}\right]$ as:

$$\hat{\boldsymbol{\alpha}}_{B_{i+1}}^{B_k} = \hat{\boldsymbol{\alpha}}_{B_i}^{B_k} + \hat{\boldsymbol{\beta}}_{B_i}^{B_k}\delta t + \frac{1}{2}\boldsymbol{R}(\hat{\boldsymbol{\gamma}}_{B_i}^{B_k})(\hat{\mathbf{a}}_i - \hat{\mathbf{b}}_{a_i})\delta t^2$$

$$\hat{\boldsymbol{\beta}}_{B_{i+1}}^{B_k} = \hat{\boldsymbol{\beta}}_{B_i}^{B_k} + \boldsymbol{R}(\hat{\boldsymbol{\gamma}}_{B_i}^{B_k})(\hat{\mathbf{a}}_i - \hat{\mathbf{b}}_{a_i})\delta t$$

$$\hat{\boldsymbol{\gamma}}_{B_{i+1}}^{B_k} = \hat{\boldsymbol{\gamma}}_{B_i}^{B_k} \otimes \begin{bmatrix} 1 \\ \frac{1}{2}(\hat{\boldsymbol{\omega}}_i - \hat{\mathbf{b}}_{\omega_i})\delta t \end{bmatrix}$$

$$\hat{\boldsymbol{\phi}}_{B_{i+1}}^{B_k} = \hat{\boldsymbol{\phi}}_{B_i}^{B_k} + \boldsymbol{R}\left(\hat{\boldsymbol{\gamma}}_{B_i}^{B_k}\right)\boldsymbol{R}_{O_i}^{B_i}\hat{\boldsymbol{c}}^{O_i}\hat{\mathbf{v}}^{O_i}\delta t \quad (8)$$

Finally, the residual of preintegrated IMU/odometer data $\left[\delta\boldsymbol{\alpha}_{B_{k+1}}^{B_k}\ \delta\boldsymbol{\beta}_{B_{k+1}}^{B_k}\ \delta\boldsymbol{\theta}_{B_{k+1}}^{B_k}\ \delta\mathbf{b}_a\ \delta\mathbf{b}_g\ \delta\boldsymbol{\phi}_{B_{k+1}}^{B_k}\ \delta c^O\right]^T$ is given as:

$$r_{\mathcal{J}}(\hat{\mathbf{Z}}_{B_{k+1}}^{B_k}, \mathcal{X}) = \begin{bmatrix} \mathbf{R}_W^{B_k}\left(\mathbf{p}_{B_{k+1}}^W - \mathbf{p}_{B_k}^W + \frac{1}{2}\boldsymbol{g}^W \Delta t_k^2 - \mathbf{v}_{B_k}^W \Delta t_k\right) - \hat{\boldsymbol{\alpha}}_{B_{k+1}}^{B_k} \\ \mathbf{R}_W^{B_k}\left(\mathbf{v}_{B_{k+1}}^W + \boldsymbol{g}^W \Delta t_k - \mathbf{v}_{B_k}^W\right) - \hat{\boldsymbol{\beta}}_{B_{k+1}}^{B_k} \\ 2\left[\left(\mathbf{q}_{B_k}^W\right)^{-1} \otimes \left(\mathbf{q}_{B_{k+1}}^W\right) \otimes \left(\hat{\boldsymbol{\gamma}}_{B_{k+1}}^{B_k}\right)^{-1}\right]_{2:4} \\ \mathbf{b}_{a_{k+1}} - \mathbf{b}_{a_k} \\ \mathbf{b}_{g_{k+1}} - \mathbf{b}_{g_k} \\ \mathbf{R}_W^{B_k}\left(\mathbf{p}_{B_{k+1}}^W - \mathbf{p}_{B_k}^W + \mathbf{R}_{B_{k+1}}^W \mathbf{p}_{O_{k+1}}^{B_{k+1}}\right) - \hat{\boldsymbol{\phi}}_{B_{k+1}}^{B_k} \\ \boldsymbol{c}^{O_{k+1}} - \boldsymbol{c}^{O_k} \end{bmatrix} \quad (9)$$

where $[\cdot]_{2:4}$ is used to take out the last four elements from a quaternion. $\mathbf{p}_{O_{k+1}}^{B_{k+1}}$ is the displacement between odometer and the IMU measured by a total station.

*B. LiDAR-Related Factors*

Since the range measuring error in the axial direction is large for short-distance, we first remove the too close points from LiDAR. Then we apply the IMU/odometer increment model to correct LiDAR point motion distortion with linear interpolation.

We follow the work of [21] to extract two sets of feature points from denoised and distortion-free point cloud. The edge features $\varepsilon$ are selected with high curvature and the planar features $\rho$ are with low curvature. Then we take the fused feature points to perform scan registration with the edge and planar patch correspondence computed through point-to-line and point-to-plane distances:

$$d_{\varepsilon_k} = \frac{\left|\left(\mathbf{p}_k^W - \boldsymbol{\varepsilon}_1^{L_W}\right) \times \left(\mathbf{p}_k^W - \boldsymbol{\varepsilon}_2^{L_W}\right)\right|}{\left|\boldsymbol{\varepsilon}_1^{L_W} - \boldsymbol{\varepsilon}_2^{L_W}\right|} \quad (10)$$

$$d_{\rho_k} = \frac{\left(\mathbf{p}_k^W - \boldsymbol{\rho}_1^{L_W}\right)^T \left(\left(\boldsymbol{\rho}_1^{L_W} - \boldsymbol{\rho}_2^{L_W}\right) \times \left(\boldsymbol{\rho}_1^{L_W} - \boldsymbol{\rho}_3^{L_W}\right)\right)}{\left|\left(\boldsymbol{\rho}_1^{L_W} - \boldsymbol{\rho}_2^{L_W}\right) \times \left(\boldsymbol{\rho}_1^{L_W} - \boldsymbol{\rho}_3^{L_W}\right)\right|} \quad (11)$$

here $\mathbf{p}_k^W$ represents the scan point in the global frame. $(\boldsymbol{\varepsilon}_1^{L_W}, \boldsymbol{\varepsilon}_2^{L_W})$ and $(\boldsymbol{\rho}_1^{L_W}, \boldsymbol{\rho}_2^{L_W}, \boldsymbol{\rho}_3^{L_W})$ are from the 5 nearest points of a current edge or planar feature point sets in the global frame. Suppose the number of edge and planar correspondences is $N_\varepsilon$ and $N_\rho$ in the current frame, the residual can be calculated using:

$$r_{\mathcal{L}_k} = \sum_{i=1}^{N_\varepsilon} (d_{\varepsilon_i})^2 + \sum_{j=1}^{N_\rho} (d_{\rho_j})^2 \quad (12)$$

We notice that the LiDAR-only odometry with LiDAR of limited FoV is over-sensitive to the vibrations caused by the joint of rail tracks and the rail track turnouts, where errors may appear in the pitch direction. Besides, the two rail tracks are not of the same height at turnings, and the LiDAR-only odometry will keep this roll divergence even in the following straight railways. Illustrated in [22], the planar features from segmented ground can effectually constrain the roll and pitch rotation. However, the angle-based ground extraction is not robust for railways as the small height variations will be ignored by the segmentation, which will generate large vertical divergence for large-scale mapping tasks.

We hereby employ the rail track plane to provide ground constraints. We first detect the track bed area using the LiDAR sensor mounting height and angle. With the assumption of the LiDAR is centered between two rail tracks, we can set two candidate areas around the left and right rail tracks and search the points with local maximum height over the track bed. Two straight lines can then be fixed using random sample consensus (RANSAC) [23] method. Finally, we exploit the idea of region growing [24] for further refinement. As a prevailing segmentation algorithm, region growing examines neighboring points of initial seed area and decides whether to add the point to the seed region or not. We set the initial seed area within the distance of 5 m ahead of the LiDAR, and the distance threshold of the search region to the fitted line is set to 0.07 m, which is the width of the track head.

We are now able to define a plane with the two sets of rail track points using RANSAC. And the ground plane $\boldsymbol{m}$ can be parameterized by the normal direction vector $\boldsymbol{n}_p$ and a distance scalar $d_p$, $\boldsymbol{m} = [\boldsymbol{n}_p^T, d_p]^T$. Then the correspondence of each ground point between two consecutive scan $k$ and $k + 1$ can be established by:

$$\mathbf{p}^{L_{k+1}} = \mathbf{T}_{L_k}^{L_{k+1}} \mathbf{p}^{L_k} \quad (13)$$

$$\mathbf{T}_{L_k}^{L_{k+1}} = \mathbf{T}_B^L \mathbf{T}_{B_k}^{B_{k+1}} \quad (14)$$

where $\mathbf{p}^{L_{k+1}}$ and $\mathbf{p}^{L_k}$ is the same point expressed in frame $L_{k+1}$ and $L_k$ with the corresponding transformation defined by $\mathbf{T}_{L_k}^{L_{k+1}} = \left\{\mathbf{R}_{L_k}^{L_{k+1}}, \mathbf{p}_{L_k}^{L_{k+1}}\right\}$. Based thereon, the ground plane measurement residual can be expressed as:

$$r_{\mathcal{P}_k} = \boldsymbol{m}_{k+1} - \mathbf{T}_{L_{k+1}}^{L_k} \boldsymbol{m}_k \quad (15)$$

*C. Visual-Related Factors*

Different from UAVs or UGVs with arbitrary movements, the rail vehicles are limited to highly-constrained motions, such as long-time consistent velocities, constant linear accelerations at accelerate or braking stage, and no rotations. According to [25], moving with constant acceleration or without rotating can introduce locally unobservable IMU biases, generating ill-conditioned or even rank-deficient information matrix and result in significant scale drift for VINS. Therefore, we leverage the IMU bias estimated by LiDAR-inertial odometry at the initialization stage, and associate relative depth information to extracted feature points on image keyframes. The scale drift can then be corrected by odometer aided VINS and the scale correction optimization in [17].

Since more scale correction is added, the $x - y$ translational accuracy can be significantly improved. However, the rotational error is still not evitable, and we introduce two additional structure constraints for further refinement.

We can extract the line segments with LSD [26] as shown in Fig. 3. Assuming $\mathbf{l} = [l_1, l_2, l_3]^T$ is the projection of a line feature on the image plane [27], and $\underline{\mathbf{m}}$ as the midpoint of the line feature, the line feature reprojection error can be defined as:

$$r_{S_k} = \frac{\underline{\mathbf{m}}^\mathrm{T} \mathbf{l}}{\sqrt{l_1^2 + l_2^2}} \quad (16)$$

Besides, the rotation drift can be effectually constrained with detected vanishing point as illustrated in [28]. We hereby utilize the vanishing point to constrain the camera orientation. Suppose $\mathbf{n}_k^C$ is the attitude of the detected vanishing point in the $k$-th frame, and the corresponding vanishing point error can be expressed following [18]:

$$r_{V_k} = \mathbf{n}_k^C \times \left(\mathbf{R}_W^{C_k} \mathbf{n}_k^W\right) \quad (17)$$

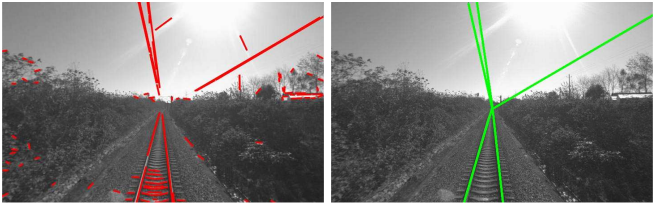

**Fig. 3.** Extracted LSD (left) and vanishing point (right) from a single frame.

### D. GNSS Factors

The accumulated drifts of the system can be corrected using RTK measurements. The GNSS factor is added when the estimated position covariance is larger than the reported GNSS covariance in [29]. However, we find that the reported GNSS covariance is not trustworthy sometimes, and may yield blurred or inconsequent mapping result. We hereby model the GNSS measurements $\mathbf{p}_k^W$ with additive noise, and the global position residual can be defined as:

$$r_{G_k} = \mathbf{R}_W^{B_k}(\mathbf{p}^{W_k} - \mathbf{p}_W^B - \mathbf{p}_{B_k}^W \\ + \frac{1}{2}\mathbf{g}^W \Delta t_k^2 - \mathbf{v}_{B_k}^W \Delta t_k) - \widehat{\boldsymbol{\alpha}}_{B_{k+1}}^{B_k} \quad (16)$$

where $\mathbf{p}_W^B$ is the transformation from the receiver antenna to the IMU, which can be obtained from installation configuration. Note that we only consider the single point positioning (SPP) result as input due to the inconsistent 4G communication quality for long railroads.

### E. Map Management

To reduce the mapping blurry caused by frequent GNSS aided optimization, We propose a submap-based two-stage map-to-map registration, which first creates submaps based on local optimization, then leverage the GNSS information for submap-to-submap registration using the normal distribution transform (NDT) [30]. In practice, 30 keyframes are maintained in each submap, which can reduce the mapping blurry caused by frequent correction.

---

[1] https://www.xsens.com/mti-680g
[2] http://www.whmpst.com/en/imgproduct.php?aid=29

## V. Experiments

### A. Hardware Setup

We conduct a series of experiments with various kinds of maintenance vehicles on two railroads: one is a freight traffic railway for general-speed trains, the other is a manned traffic railway for high-speed trains. According to the safety principle on the railroad, we can only carry out the former experiments in the daytime, and the latter in the midnight.

The overview of the system installment and hardware setup is shown in Fig. 4, including a Hikvision camera, a Livox Horizon LiDAR, and a MTi-680G [1] integrated navigation system. Additionally, the system also incorporates a wheel odometer. All the sensors are hardware-synchronized with a u-blox EVK-M8T GNSS timing evaluation kit using GNSS pulse per second. Besides, we employ an onboard computer, with i9-10980HK CPU (2.4 GHz, octa-core), 64GB RAM, for real-time processing. All our algorithms are implemented in C++ and executed in Ubuntu Linux using the ROS [31]

The details of the four datasets are listed in TABLE II. And the ground truth is kept by the post processing result of a MPSTNAV M39 GNSS/INS integrated navigation system [2] (with RTK corrections sent from Qianxun SI).

We evaluate the proposed framework with R2LIVE, FAST-LIO2, LiLi-OM, Lio-Livox [3], and VINS-Mono. For ablation study, we define RailLoMer-V w/o GNSS, RailLoMer-V w/o ODO, and RM-LVI as no GNSS factor, no odometer factor, and the LiDAR-visual-inertial part of our proposed system.

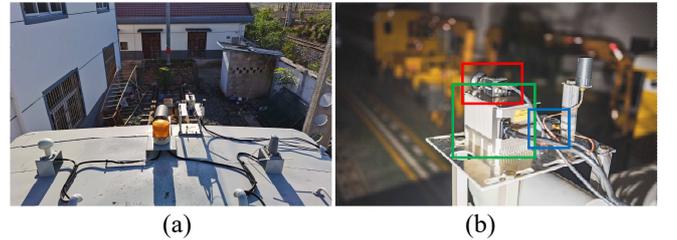

(a)                                  (b)

**Fig. 4.** Illustration of two kinds of installment based on different types of maintenance vehicles. (a) is the longitudinal-way on the general-speed railway maintenance vehicle. (b) is the lateral-way on the high-speed railway maintenance vehicle. And the red, green, blue box indicates Hikvision camera, Livox Horizon LiDAR, and MTi-680G, respectively.

TABLE II
DETAILS OF ALL THE SEQUENCES FOR EVALUATION

| Name | Duration (sec) | Distance (km) |
|---|---|---|
| General-speed railway sequence gathered at daytime | | |
| *HQ-Long* | 1325 | 25.2 |
| *HQ-Short* | 423 | 6.3 |
| High-speed railway sequence gathered at midnight | | |
| *CH-Long* | 715 | 6.5 |
| *CH-Short* | 336 | 2.4 |
| *CH-Tunnel* | 475 | 4.2 |

[3] https://github.com/Livox-SDK/LIO-Livox

TABLE III
ACCURACY EVALUATION FOR ALL THE SEQUENCES

| Absolute Trajectory Error (ATE) RMSE [m] / MAX [m], with - and indicates meaningless result. | | | | | |
|---|---|---|---|---|---|
| | *HQ-Long* | *HQ-Short* | *CH-Long* | *CH-Short* | *CH-Tunnel* |
| R2LIVE | - / - | 6.36 / 32.68 | 4.35 / 7.47 | 2.32 / 5.53 | 27.59 / 79.81 |
| FAST-LIO2 | - / - | 8.53 / 29.73 | 5.56 / 8.64 | 4.06 / 9.7 | 39.68 / 107.54 |
| LiLi-OM | 447.5 / 1896.93 | 7.95 / 31.5 | 5.23 / 9.67 | 3.51 / 5.17 | 18.57 / 64.21 |
| Lio-Livox | - / - | 18.97 / 48.46 | 7.58 / 12.77 | 5.42 / 6.66 | 32.68 / 98.57 |
| VINS-Mono | - / - | 142.47 / 288.6 | 145.7 / 293.63 | 17.63 / 35.4 | - / - |
| RailLoMer-V | **0.36 / 1.8** | **0.35 / 0.84** | **0.49 / 1.43** | **0.45 / 1.47** | **0.57 / 2.95** |
| RailLoMer-V w/o GNSS | 299.58 / 866.57 | 2.35 / 4.53 | 2.1 / 3.52 | 0.96 / 1.77 | 3.67 / 6.55 |
| RailLoMer-V w/o ODO | 1.64 / 5.41 | 1.47 / 2.11 | 0.73 / 1.71 | 0.47 / 1.82 | 8.08 / 39.49 |
| RM-LVI | 349.65 / 1078.5 | 2.48 / 5.56 | 2.37 / 4.72 | 1.24 / 3.8 | 11.33 / 47.68 |

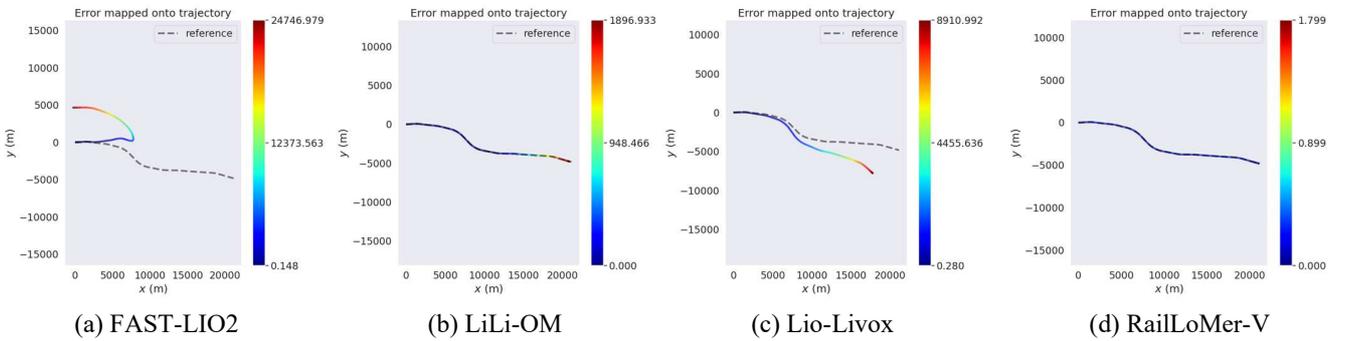

(a) FAST-LIO2  (b) LiLi-OM  (c) Lio-Livox  (d) RailLoMer-V
**Fig. 5.** The trajectory evaluation on *HQ-Long* for various methods.

## B. Evaluation

We now present a series of evaluation to qualitatively and quantitatively analyze our proposed framework, and TABLE III summarizes the root mean square error (RMSE) and maximum positioning error (MAX) metrics of various methods.

1) *Benchmarking Results*: Our system can achieve decimeter grade accuracy for all the sequences. However, the performances of other approaches are not admirable. For the small-scale test, the incorrectly initialized gravity vector accelerates the vertical error accumulation. The maintenance vehicle starts at a big turning in *CH-Short*. As the two rail tracks are not of the same height at turning, the gravity vector cannot be initialized correctly, and the selected methods all work not properly without additional constraints. For the large-scale test, the vision-only method shows terrible performance with constant velocity. This highly-constrained motion results in insufficient axis excitement, and generates great scale drift.

2) *Robustness Towards Direction of Motion*: Different from unmanned ground vehicles (UGV) or automated vehicles, which are always moving forward while data gathering, the rail vehicles also comprise a long-time backward motion. And we employ *HQ-Long*, a backward motion only sequence, to study the effect of motion direction. Presented in TABLE III and Fig. 5, the filter-based algorithm R2LIVE and FAST-LIO2 see an enormous divergence in 6-DoF on this sequence. However, the optimization-based algorithms are insensitive to this influence.

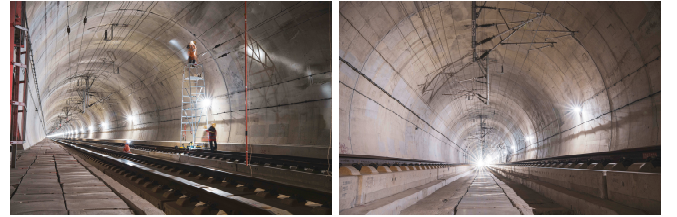

**Fig. 6.** Visual illustration of the repetitive structures in the high-speed railway tunnel.

3) *Robustness Towards Degeneracy*: Shown in Fig. 6, the tunnels of high-speed railway are with smooth man-made walls, repetitive rail tracks and suspension clamps. We believe these districts as one of the most difficult scenarios for SLAM, and we challenge our system with *CH-Tunnel*, comprised of three consecutive tunnels (with the longest 1.7 km long). With the assistance of odometer and GNSS, the longitudinal divergence and accumulated errors can be well-eliminated, and our proposed system can maintain an accurate trajectory. On the contrary, the other methods either 'stops' or 'moves backward' towards grievously degenerated scenarios.

4) *Weather Influence*: The general-speed railway datasets are collected in the summer. whereas the high-speed railway datasets are gathered in the winter. Since the rail tracks become wet and slippery in the winter midnight, the wheel slip is unavoidable, and the proposed RailLoMer-V system has a relatively low performance in the winter.

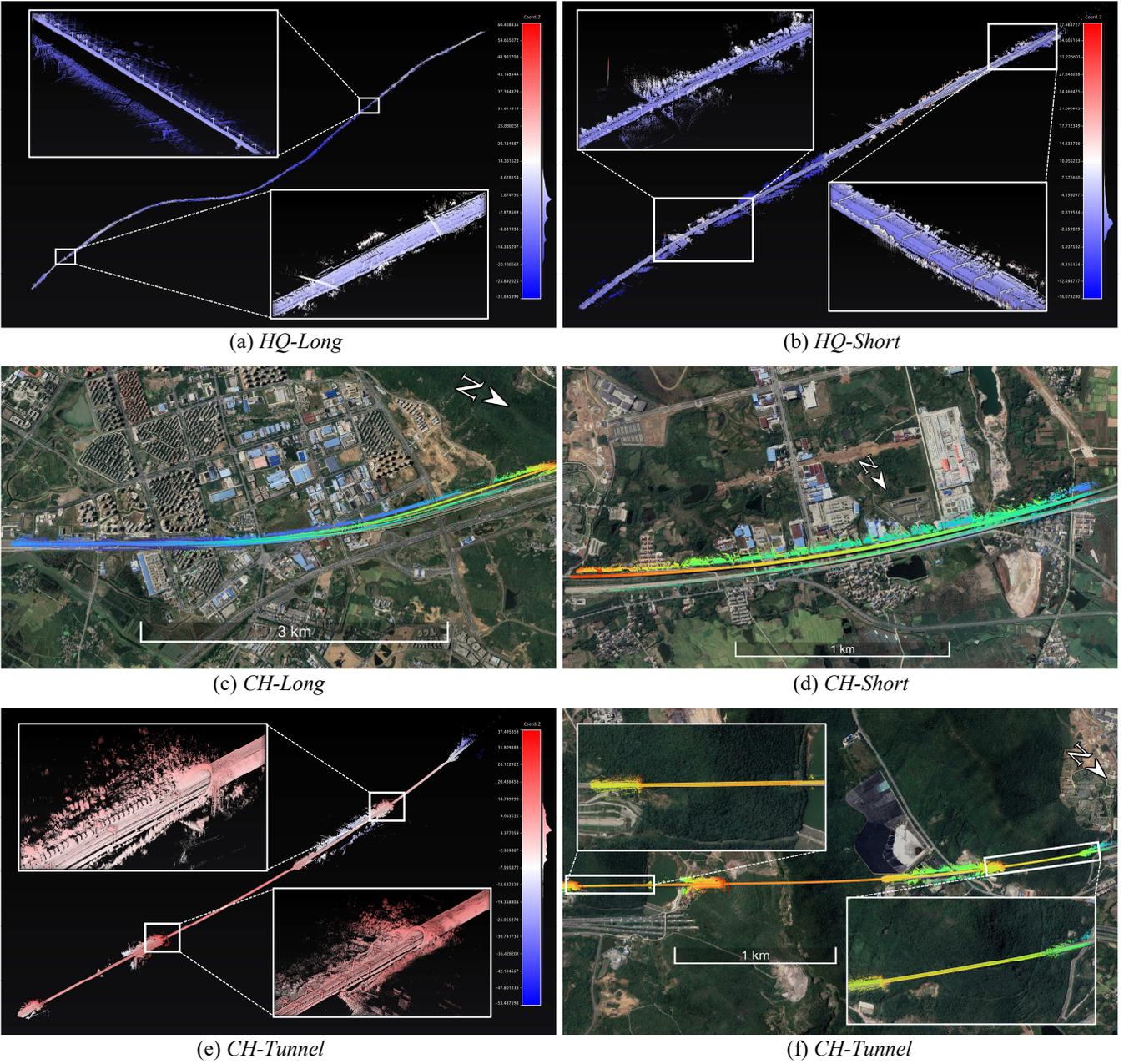

**Fig. 7.** The real-time mapping result of each sequence. (a) and (b) are from the general-speed railway experiment, with the color coded by height variations. (c) and (d) are the high-speed railway experiment mapping aligned with Google Map. (e) and (f) presents the novelty of tunnel experiment.

5) *Ablation Study*: It is seen that the odometer can well constrain the longitudinal displacement at degenerated areas, and the RailLoMer-V w/o GNSS has a better performance than RailLoMer-V w/o ODO in *CH-Tunnel*. However, the odometer contribution is not evident for feature-rich districts, and the RM-LVI sees a similar accuracy with RailLoMer-V w/o GNSS for the other sequences.

6) *High Precision Map Construction*: We show that our proposed method is accurate enough to build large-scale map of railroad environments. The real-time mapping is shown in Fig. 7, in which the clear and well-matched result with the satellite image indicates that our proposed method is of high precision.

TABLE IV
THE AVERAGE TIME CONSUMPTION IN MS

|  | **OLIS** | **OVIS** | **Fusion** |
|---|---|---|---|
| *HQ-Long* | 33.7 | 34.1 | 52.5 |
| *HQ-Short* | 28.8 | 31.6 | 37.6 |
| *CH-Long* | 30.3 | 33.2 | 45.2 |
| *CH-Short* | 29.5 | 31.5 | 39.6 |
| *CH-Tunnel* | 27.4 | 30.7 | 39.4 |

*C. Runtime Analysis*

The average runtime for processing each scan in different scenarios is shown in TABLE IV, denoting the proposed system capable of real-time operation for all conditions. Besides, the time has not seen a large growth with increased distance.

## VI. CONCLUSION

We have proposed RailLoMer-V, an accurate and robust localization and mapping framework for rail vehicles in this paper. Our system fuses measurements from LiDAR, camera, IMU, train odometer, and GNSS with a tightly-coupled manner. Besides, we leverage additional geometric structure constraints to cope with the highly-repeated environments. Our proposed system shows decimeter grade positioning accuracy through evaluations over various illumination conditions, different scales, and degenerated districts.

We hope that our experimental work and evaluation could inspire follow-up works to explore more intelligent systems for railroad applications, especially for facility and environment monitoring. From our four-year experience staying with various railroad maintenance vehicles, a small advance towards automation will save tremendous amount of manpower for construction and maintenance.

ACKNOWLEDGMENT

We would like to thanks colleagues from Hefei power supply section, China Railway, for their kind support.